\documentclass[11pt]{article}

\usepackage[letterpaper,margin=1in]{geometry} 

\usepackage{microtype}
\usepackage{graphicx}
\usepackage{subcaption}
\usepackage{booktabs} 

\usepackage{hyperref}

\usepackage[backend=biber,
            style=numeric,
            sorting=none,
            maxnames=99,
            maxbibnames=99]{biblatex}
\usepackage{amsmath}
\usepackage{amssymb}
\usepackage{mathtools}
\usepackage{amsthm}
\usepackage{algorithmic}
\usepackage{algorithm}
\usepackage{authblk}

\usepackage[capitalize,noabbrev]{cleveref}

\theoremstyle{plain}
\newtheorem{theorem}{Theorem}[section]

\newtheorem{lemma}[theorem]{Lemma}

\theoremstyle{definition}
\newtheorem{definition}[theorem]{Definition}

\theoremstyle{remark}

\usepackage[textsize=tiny]{todonotes}

\usepackage{graphicx}
\usepackage{textcomp}
\usepackage{colortbl} 
                                 
\usepackage{amsthm}
\usepackage{graphics} 
\usepackage{epsfig} 
\usepackage{times} 
\usepackage{bm}
\usepackage{xcolor}
\usepackage{booktabs}
\usepackage{siunitx}

\usepackage{caption}
\usepackage{lipsum}
\usepackage{multirow}
\usepackage{diagbox} 
\usepackage{subcaption}
\usepackage{adjustbox} 
\usepackage{makecell}

\newcommand{\R}{\mathbb{R}}

\newcommand{\Pro}{\mathbb{P}}
\newcommand{\E}{\mathbb{E}}

\newcommand{\qcb}{Q_{\text{\textup{CB}}}}

\title{Distributionally Robust Multi-Agent Reinforcement Learning for \\ Dynamic Chute Mapping}

\author{
Guangyi Liu\textsuperscript{1}, 
Suzan Iloglu\textsuperscript{1}, 
Michael Caldara\textsuperscript{1}, 
Joseph W. Durham\textsuperscript{1}, 
and Michael M. Zavlanos\textsuperscript{1,2}
}

\date{} 

\begin{document}

\maketitle

\footnotetext[1]{G. Liu, S. Iloglu, M. Caldara, J. W. Durham, and M. M. Zavlanos are with Amazon Robotics \texttt{\{gyliu,siloglu,caldaram,josepdur,miczavla\}@amazon.com}.}
\footnotetext[2]{M. M. Zavlanos is with Department of Mechanical Engineering and Materials Science, Duke University. \texttt{\{michael.zavlanos\}@duke.edu.}}

\begin{abstract}
In Amazon robotic warehouses, the destination-to-chute mapping problem is crucial for efficient package sorting. Often, however, this problem is complicated by uncertain and dynamic package induction rates, which can lead to increased package recirculation. To tackle this challenge, we introduce a Distributionally Robust Multi-Agent Reinforcement Learning (DRMARL) framework that learns a destination-to-chute mapping policy that is resilient to adversarial variations in induction rates. Specifically, DRMARL relies on group distributionally robust optimization (DRO) to learn a policy that performs well not only on average but also on each individual subpopulation of induction rates within the group that capture, for example, different seasonality or operation modes of the system. This approach is then combined with a novel contextual bandit-based predictor of the worst-case induction distribution for each state-action pair, significantly reducing the cost of exploration and thereby increasing the learning efficiency and scalability of our framework. Extensive simulations demonstrate that DRMARL achieves robust chute mapping in the presence of varying induction distributions, reducing package recirculation by an average of 80\% in the simulation scenario.
\end{abstract}

\section{Introduction}

\begin{figure}[t]
    \vskip 0.2in
    \centering
    \includegraphics[width=0.7\linewidth]{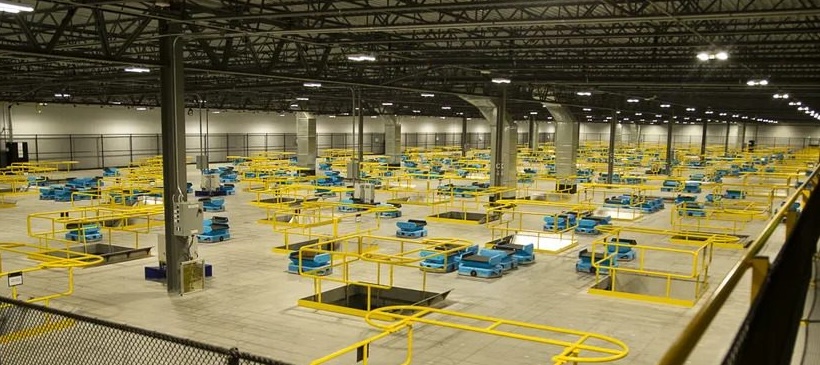}  
    \caption{Schematic layout of an Amazon robotic sortation warehouse featuring eject chutes.}
    \label{fig:sort_floor}
    \vskip -0.2in
\end{figure}

In Amazon robotic sortation warehouses, mobile robots are deployed to transport and sort packages efficiently to different destinations \cite{wurman2008coordinating, azadeh2019robotized, amazon_sort_floor, amazon_2022, amazontor_2022}. The sorting process begins at induction stations, where packages are loaded onto mobile robots and subsequently transported to designated eject chutes based on their destinations (\cref{fig:sort_floor}). A critical factor determining the package throughput capacity of these facilities is the effective allocation of eject chutes to different destinations. Therefore, the destination-to-chute mapping policy plays a crucial role in optimizing the overall throughput performance of the robotic sortation warehouse.

Our previous work \cite{shen2023multi} addresses the destination assignment problem (DAP) \cite{boysen2010cross} in robotic sorting systems by developing a dynamic chute mapping policy. This policy determines the optimal allocation of eject chutes to destinations with the objective of minimizing the number of unsorted packages. We proposed a model-free reinforcement learning approach that dynamically adjusts the number of chutes assigned to each destination throughout the day. Our solution formulates the chute mapping problem within a Multi-Agent Reinforcement Learning (MARL) framework \cite{lowe2017multi,sunehag2017value,samvelyan2019starcraft,rashid2020monotonic}, where each destination is represented as an agent that controls its chute allocation at each time step.

While the MARL policy proposed in our previous work \cite{shen2023multi} demonstrates superior performance compared to traditional reactive chute mapping approaches often implemented in Amazon robotic sortation warehouses, its effectiveness assumes the induction distribution during deployment matches the training distribution and the daily induction rate stays close to its average value. In practice, however, induction patterns exhibit significant temporal variations, potentially compromising the MARL policy's performance when confronted with unexpected distribution changes.

To enhance robustness against such variations, in this paper, we propose a Distributionally Robust Multi-Agent Reinforcement Learning (DRMARL) framework that develops chute mapping policies capable of maintaining near-optimal performance across diverse induction distributions. Specifically, we formulate this problem as a group DRO problem, where each group represents a distinct induction distribution pattern extracted from a subset of a historical dataset. Our DRMARL framework optimizes policies for the worst-case induction patterns across these distribution groups. To address the computational cost of exhaustively evaluating all distribution groups during training, we introduce a contextual bandit (CB)-based worst-case reward predictor for each state-action pair. Through extensive evaluation, we demonstrate that our DRMARL framework yields robust chute mapping policies that not only outperform baseline MARL policies on out-of-distribution (OOD) induction data but also maintain consistent performance across varying induction distributions.

Our {\it contributions} are twofold: First, we introduce group distributionally robust optimization in multi-agent reinforcement learning, developing a novel framework to learn policies that are robust to distribution shifts. Second, we propose an innovative contextual bandit-based method for efficient worst-case reward prediction, significantly reducing the computational complexity of DRMARL training by eliminating the need for exhaustive group exploration to learn the worst-case reward. To the best of our knowledge, our framework is the first to integrate contextual bandits with group DRO and MARL, addressing a well-known challenge of distributionally robust reinforcement learning related to its computational cost. Our proposed framework has broad applicability to various large-scale industrial applications, beyond sortation systems, including resource allocation, collaborative robotics, and warehouse automation, where robustness to distribution shifts is crucial.

\subsection{Literature Review}

\noindent{\it Destination Assignment Problems}: Mathematical programming has been used to optimize warehouse systems, including destination assignment problems (DAP) \cite{boysen2010cross} for sorting systems. The destination mapping approach in \cite{fedtke2017layout} optimizes package flow by minimizing travel distances between inbound and outbound stations in conveyor-based sorting systems, leading to improved throughput. \cite{novoa2018flow} minimizes the worst-case flow imbalance across work stations on the sortation floor, developing a stochastic approach with chance and robust constraints. For robotic sorting systems, \cite{khir2021two} proposes an integer programming method to solve DAPs that minimize sortation effort and satisfy package deadlines. A robust formulation addressing demand uncertainty is presented in \cite{khir2022robust}. While these approaches effectively optimize destination assignment in sorting systems, they do not account for distributional uncertainty in demand and system dynamics. In contrast, our proposed DRMARL framework explicitly models such uncertainties, ensuring robust performance under varying operational conditions.

\vspace{1mm}

\noindent{\it MARL for Resource Allocation}: MARL has previously been applied to address resource allocation problems \cite{nie2021multi,mei2024multi,jun2025multi}. For example, a MARL framework for ocean transportation networks was proposed in \cite{li2019cooperative}. This framework develops a multi-agent $Q$-learning algorithm where the local $Q$-networks depend on the joint states (including the limited shared resources) and the joint actions. However, since the joint state-action space grows exponentially with the number of agents, the local $Q$-networks are hard to learn and this approach does not scale well in practice. This limitation was addressed in our previous work \cite{shen2023multi}, where the local $Q$-networks are only loosely coupled, enhancing the scalability while still being interconnected enough to capture the impact of robot congestion on the sortation floor. Compared to \cite{li2019cooperative}, the method proposed in \cite{shen2023multi} models resources explicitly as actions and considers budget constraints when taking joint actions. However, these MARL-based approaches do not incorporate distributional robustness, making them sensitive to demand fluctuations and uncertainty, which our DRMARL framework explicitly addresses to ensure reliable performance in dynamic sorting environments.

\vspace{1mm}

\noindent{\it Robust and Distributionally Robust RL}: Robust Reinforcement Learning (Robust RL) \cite{morimoto2005robust,wiesemann2013robust,pinto2017robust,moos2022robust,goyal2023robust,yamagata2024safe} develops policies that maintain performance under worst-case conditions through adversarial perturbations. Distributionally Robust Reinforcement Learning (DRRL) \cite{xu2010distributionally,smirnova2019distributionally,hou2020robust,wang2023foundation,ramesh2024distributionally,zhang2024distributionally,lu2024distributionally} extends this by optimizing across environment distributions rather than single worst-case scenarios. While traditional DRRL primarily addresses ambiguity in MDP transition probabilities, this approach inadequately captures induction distribution changes in Amazon robotic sortation warehouses. Our problem requires focus on distributionally robust optimization of reward function distributions, building on \cite{ren2022distributionally,liu2022distributionally}. Recent advances in (Distributionally) Robust Multi-Agent RL \cite{zhang2020robust,bukharin2024robust,shi2024sample,shi2024breaking} have introduced frameworks like RMGs, ERNIE, and DRNVI to address environmental uncertainties, adversarial dynamics, and model uncertainties. While existing methods primarily focus on robustness in transition dynamics, adversarial interactions, and general environmental uncertainties, they do not explicitly address distributional shifts in package induction, which is a critical challenge in sortation warehouses. Our approach extends DRMARL to explicitly model and optimize against uncertainties in induction distributions, ensuring robust and consistent performance under varying operational conditions.

\vspace{1mm}

\noindent{\it Group DRO:} Group Distributionally Robust Optimization aims to enhance model robustness across diverse subpopulations by optimizing for the worst-performing groups rather than average performance~\cite{sagawa2019distributionally}. This approach ensures fairness and resilience to distribution shifts, particularly for underrepresented groups. While initial work focused on single-agent supervised learning~\cite{hu2018does, oren2019distributionally}, recent advances have extended these principles to more complex settings. Notably,~\cite{soma2022optimal} provided a soft-weighting method on distribution groups with convergence guarantees, while~\cite{wu2023distributed} and~\cite{xu2023group} demonstrated the applicability of group DRO in multi-agent systems and reinforcement learning, respectively. Our work bridges a critical gap by introducing group DRO principles to DRMARL. We begin by formulating the distributionally robust Bellman operator and addressing the computational challenges of exploring all distribution groups during the training. To tackle these challenges, we provide a DR Bellman operator specifically designed for MARL and introduce a contextual bandit (CB)-based worst-case distribution group predictor. This predictor enables efficient training by adaptively identifying the worst-case distribution groups.

The remainder of the paper is organized as follows. In \cref{sec:problem_formulation}, we formulate the dynamic chute mapping problem within a multi-agent reinforcement learning framework. In \cref{sec:drmarl} we extend this formulation by incorporating group distributionally robust optimization into MARL, while in \cref{sec:cb} we present our novel contextual bandit-based worst-case reward predictor to enhance training efficiency. Finally, in \cref{sec:simulation}, we demonstrate the effectiveness of our proposed framework through extensive simulations. All theoretical proofs are provided in Appendix~\ref{app:proofs}.

\section{Problem Formulation}   \label{sec:problem_formulation}
\begin{figure}[t]
    \vskip 0.2in
    \centering
    \includegraphics[width=0.7\linewidth]{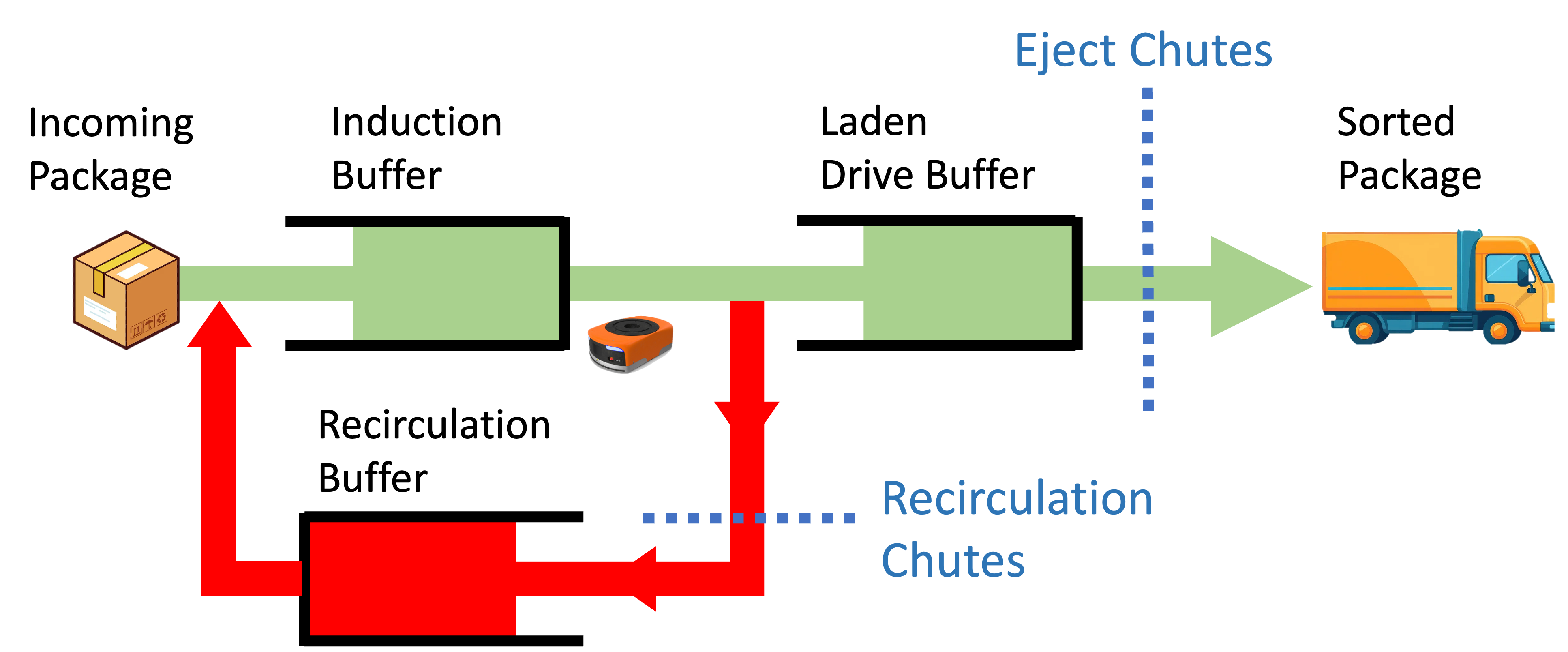}  
    \caption{Flow of packages in the Amazon robotic sortation warehouse.}
    \label{fig:sort_floor_model}
    \vskip -0.2in
\end{figure}

In Amazon robotic sortation warehouses, package flow is modeled using three buffers: induct, laden drive, and recirculation, as illustrated in \cref{fig:sort_floor_model}. The laden drive buffer represents packages currently being transported by robots to eject chutes. Due to the limited chute capacity, insufficient chute allocation for a destination can lead to sortation bottlenecks. When this occurs, excess packages from the laden drive buffer are redirected by robots to recirculation chutes, entering the recirculation buffer for reprocessing through the sortation system. Eject chutes can be reallocated under two conditions: when the destination vehicle reaches capacity or when the transportation schedule deadline is met. The dynamic chute mapping policy aims to optimize chute allocation across destinations, minimizing recirculation buffer volume while maximizing the system throughput.

\subsection{MARL Structure}    

We formulate the dynamic chute mapping problem as a sequential decision making problem; specifically, a MARL problem that determines optimal chute allocations to minimize package recirculation at each time step. To do so, we define a Markov game over $N$ agents (unique destinations) by the tuple 

\noindent$\left(N,\mathcal{S},\{\mathcal{O}^i\}_{i=1}^N,\{\mathcal{A}^i\}_{i=1}^N,P,\{r^i\}_{i=1}^N,\gamma,\rho_0 \right)$, where: 

\begin{itemize}
\item[(a)] \textbf{Agents:} $N$ agents, each corresponding to a unique destination.

\item[(b)] \textbf{State Space:} $\mathcal{S}$ denotes the joint state space. 

\item[(c)] \textbf{Observation Space:} For each agent $i$, $\mathcal{O}^i\subset \mathcal{S}$ represents its local observation at each time step, consisting of:
\begin{itemize}
    \item Number of packages recirculated until time $t$ for agent $i$
    \item Total number of available chutes that can be assigned (uniform across all agents)
    \item Number of chutes currently assigned to agent $i$
\end{itemize}

\item[(d)] \textbf{Action Space:} For each agent $i$, $\mathcal{A}^i$ represents its action space, determining the number of new chutes required. Actions take values in $[0,10]$, where an action value $a$ indicates the assignment of $a$ new chutes to agent $i$ at that time step. The joint action space is defined as $\mathcal{A}=\prod_{i=1}^N \mathcal{A}^i$.

\item[(e)] \textbf{Transition Probability:} $P:\mathcal{S}\times \mathcal{A}\times \mathcal{S} \times \mathcal{X} \to [0,1]$ specifies the probability of packages being either sorted by chutes or sent to the recirculation buffer. In the large-scale Amazon robotic sortation warehouse setting, the transition probability is a function of the induction distribution $\Pro$ with random variable $X \in \mathcal{X}$, which is addressed in detail in Appendix~\ref{app:dr_bo_approximate}.

\item[(f)] \textbf{Reward Function:} For each agent $i$, $r^i:\mathcal{S}\times \mathcal{A} \times \mathcal{X} \to\mathbb{R}$ defines the reward function, penalizing both the number of allocated chutes and the number of packages in the recirculation buffer. The recirculation is a function of the induction distribution, which is defined in~\cref{sec:induct_distribution}.
\end{itemize}

The model is completed with discount factor $\gamma\in(0,1)$ and initial state distribution $\rho_0$. To address the scalability of the state-action space and computational feasibility of the expectation in \eqref{eq:DQN_loss}, we employ the Value Decomposition Network (VDN) \cite{shen2023multi} with budget constraints for computing joint actions. Implementation details are provided in~\cref{sec:vdn} and~\cref{sec:mip}.

The system operates in discrete time steps, where at each step $t$, individual agents corresponding to destinations make chute allocation decisions. Each agent $i$ employs a local policy $\pi^i:\mathcal{O}^i \times \mathcal{A}^i \to [0,1]$, which maps local observations $o^i$ to probabilities over possible chute allocation actions. Each agent $i$ learns its optimal local policy $\pi^{i,*}$ by maximizing the expected discounted return 
$
    \mathbb{E}[R_t^i] = \mathbb{E} \big[\sum_{t'=t}^{\infty} \gamma^{t'-t} r^i_{t'}\big],
$
where $r^i_{t'}$ denotes the instantaneous reward at time $t'$. The expectation accounts for both the stochastic nature of the policy and the environment dynamics. This formulation naturally aligns individual agent objectives with the global goal of minimizing recirculation while maintaining efficient sortation throughput. The instantaneous reward function for agent $i$ at time step $t$ is defined as:
\begin{equation}
    r^i_t = -\texttt{recirc}^i_t - 2a^i_t,
\end{equation}
where $\texttt{recirc}^i_t \geq 0$ represents the number of packages in recirculation for destination $i$. Due to the coupled nature of agent decisions, we utilize the joint action-value function to determine optimal local policies $\pi^{i,*}$:
\begin{equation}
    Q^{\pi}(s,a) = \mathbb{E}\Big[\sum_{i=1}^N R^i_t|s_t=s,a_t=a \Big],
\end{equation}
which evaluates the expected return when taking joint action $a=(a^1,\ldots,a^N)$ in state $s$ and following joint policy $\pi$ thereafter.

To mitigate the exponentially growing policy space, we assume agents execute actions independently such that $\pi=\prod_{i=1}^N \pi^i$. The optimal policy $\pi^*$ is learned using the Deep Q-Network (DQN)~\cite{mnih2015human}, where a neural network $Q(s,a;\theta)$ with parameters $\theta$ approximates the optimal action-value function $Q^{\pi^*}(s,a)$. The learning process minimizes the loss:
\begin{equation} \label{eq:DQN_loss}
    L(\theta;X)=\mathbb{E}_{s,a,r,s'} \left[ \left(Q(s,a;\theta)-y \right)^2 \right],    
\end{equation}
where $y=\E_{X \sim \Pro} [r(s,a;X)]+\gamma\max_{a'} \bar{Q}(s',a';\bar{\theta})$ approximates the optimal target values $\E_{X \sim \Pro} [r(s,a;X)] + \gamma\max_{a'}Q^{\pi^*}(s',a')$. Here, $r(s,a;X)$ represents the instantaneous reward under the current state-action pair and induction pattern $X$.\footnote{See~\cref{sec:induct_distribution} for detailed definition of induction pattern and induction generating distribution.} Stability of the learning process is enhanced through two mechanisms: a target network $\bar{Q}$ with periodic parameter updates using the most recent values of $\theta$, and the use of an experience replay buffer $\mathcal{D}$ storing transition tuples $(s,a,r,s')$. The resulting optimal policy takes the form:
\begin{equation}
    \pi^*(s,a) = 
    \begin{cases}
        \frac{1}{|\mathcal{A}(s)|} & \text{if } a \in \mathcal{A}(s) \\
        0 & \text{otherwise}
    \end{cases},
\end{equation}
where $\mathcal{A}(s) = \arg\max_{a}Q(s,a;\theta^*)$ and $\theta^* = \arg\min L(\theta;X)$.

\subsection{Induction Distribution}     \label{sec:induct_distribution}

For a given sortation warehouse with $D$ destinations and $T$ time intervals (e.g., hours or minutes) during one day, we consider the random variable $X = \{X_1,\ldots,X_{DT}\} \in \R^{DT}$, where $X_i \geq 0$ represents the number of packages arriving at each destination-time pair $(d,t)$ during the day. We model each day's induction as a random variable $X$, generated by an unknown probability distribution $\Pro$, which we refer to as the induction generating distribution. We assume the total daily package induction volume remains constant at $V$ across all days, as the MARL chute mapping policy is primarily influenced by the distribution pattern rather than total volume variations.

\begin{definition}
    (Induction Generating Distribution) Let us consider a multinomial distribution $\mathcal{M}(n,p_1,...,p_k)$, which obtains its support on 
    \begin{equation}
        \left\{(z_1,...,z_k) ~\bigg|~ \sum_{i=1}^{k} z_i = V, z_i \geq 0, \forall i = 1,...,n\right\},
    \end{equation}
    and the probability mass function is given by
    \begin{equation}
        \Pro(X_1 = z_1, \dots, X_k = z_k) = \frac{n!}{z_1! z_2! \dots z_k!} ~ p_1^{z_1} \, p_2^{z_2} \, \dots \, p_k^{z_k}.
    \end{equation}
\end{definition}

For a daily induction random variable $X$ and induction patterns from temporally proximate dates (e.g., within the same week), we assume they follow the same induction generating distribution $\mathcal{M}(n,p_1,\ldots,p_{DT})$. The empirical induction generating distribution is estimated using Sample Average Approximation (SAA) \cite{kim2015guide} from these temporally related dates.

In practice, we utilize induction data collected from Years 1-4, clustering it into $21$ groups based on week numbers. For each group $g$, we construct an empirical induction generating distribution $\Pro_g$ via SAA, modeled as a multinomial distribution using all induction data within that group. The group ambiguity set $\mathfrak{M}$ in \eqref{eq:group_ambi} is then formed using the collection of distributions $\{\Pro_1,\Pro_2,\ldots,\Pro_{21}\}$.

The probability $p_j$ of an incoming package being assigned to the $j$'th destination-time pair is derived from the approximated empirical multinomial distribution. The distribution of $V$ packages across destinations is then determined through sampling $V$ times according to these probabilities.

\subsection{Dimension Reduction of the State-Action Space} \label{sec:vdn}

To manage the dimensionality of the state-action space, we decompose the joint $Q$-network into a sum of local $Q$-networks. Each local network captures the expected return of individual agents' chute mapping actions, while the joint network represents the expected return of the complete chute assignment across all agents. Specifically, we express the joint $Q$-network as:
\begin{align}
\label{eq:network_vdn}
    Q(s,a,\theta)=\sum\nolimits_{i=1}^N Q'(i,s^i,a^i;\theta),
\end{align}
where the input space scales linearly with the number of agents. While this decomposition is similar to \cite{shen2023multi}, our approach learns a single shared $Q'$ network for all agents instead of separate networks for each agent, resulting in improved computational efficiency.

\subsection{Feasibility of Joint Actions} \label{sec:mip}
In unconstrained settings, agents would simply select actions that maximize their individual $Q$-networks, with the joint action being the collection of these individual choices. However, the chute mapping problem introduces resource constraints, as agents must share a limited number of available chutes. This necessitates coordination to allocate resources optimally among agents based on their state-action values.

Given a budget constraint $M$ on the joint actions such that $\sum_{i=1}^N a_i \leq M$, we formulate the following integer program to determine the optimal joint action that maximizes the joint $Q$-network for any state $s$:
\begin{equation} 
    \begin{aligned}
        \underset{a^1,\cdots,a^N}{\text{maximize}} \;  &\sum\nolimits_{i=1}^N Q'(i,s^i,a^i;\theta) \\
        \text{s.t.} \; &\sum\nolimits_{i=1}^N a^i\leq M, \; a^i \in \mathbb{N}
    \end{aligned}
\end{equation}

This integer program, which can be efficiently solved using commercial solvers such as Google-OR Tools \cite{ortools} or Xpress \cite{fico_2023}, serves two purposes: it generates feasible data for the replay buffer to compute the expectation in \eqref{eq:DQN_loss} during training, and it determines the optimal actions once learning has converged. Notably, this optimization step is separate from the $Q$-learning process.

While the above MARL-based chute mapping policy demonstrates strong performance under standard operating conditions, it exhibits significant performance degradation when faced with distribution shifts in package induction patterns (see \cref{fig:OOD_degradation} in \cref{sec:simulation}). This vulnerability to out-of-distribution scenarios motivates our robust formulation.

In this paper, our {\it objective is to introduce robustness to distribution shifts} in the learned chute mapping policies. We achieve this by incorporating group DRO into the MARL framework, giving rise to the proposed DRMARL approach. Our porposed framework ensures reliable and robust performance on Amazon robotic sortation warehouses, even under unforeseen future induction scenarios.

\section{Distributionally Robust Multi-Agent Reinforcement Learning with Group DRO} \label{sec:drmarl}

In this section, we enhance the MARL chute mapping framework described in \cref{sec:problem_formulation} by incorporating group Distributionally Robust Optimization (DRO) to handle uncertainty and variability in package induction patterns. This approach enables us to develop robust policies that perform well across diverse induction scenarios, including previously unseen induction patterns.

Compared to traditional stochastic optimization that assumes fixed probability distributions, DRO~\cite{delage2010distributionally,shapiro2017distributionally,rahimian2019distributionally,zhen2023unified,liu2024data,kuhn2024distributionallyrobustoptimization, konti2024distributionally} takes a more general approach that defines an ambiguity set containing multiple plausible distributions derived from the available data. By optimizing for the worst-case scenario within this set, DRO learns policies that remain effective even when the testing distribution differs from the training conditions, as is the case in the chute mapping problem considered here.

\subsection{Group DRO}

In DRO, the ambiguity set that captures uncertainty in the data-generating distribution can be defined in various ways. Group DRO~\cite{hu2018does,oren2019distributionally} offers an efficient way of defining the ambiguity set using a finite collection of known distributions. For robotic sortation warehouses, these distributions can be derived from historical induction patterns.

\begin{figure}[t]
    \vskip 0.2in
    \centering
    \includegraphics[width = 0.8 \linewidth]{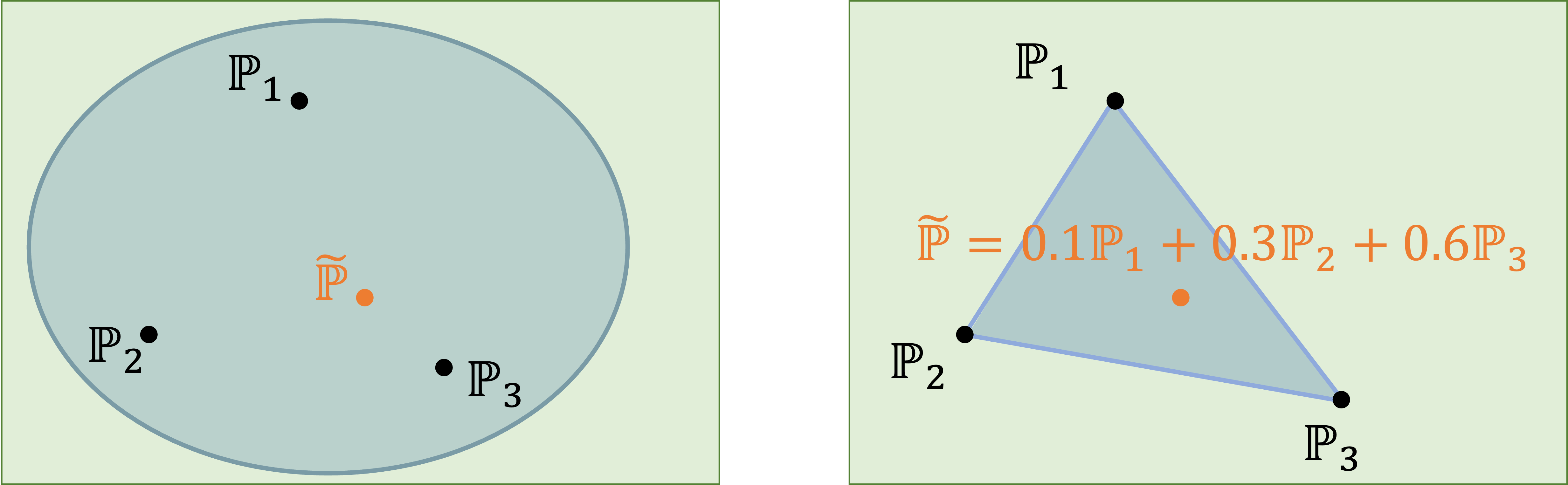}
    \caption{Ambiguity set $\mathfrak M$ in regular DRO \cite{rahimian2019distributionally} (left) versus group DRO \cite{sagawa2019distributionally} (right).}
    \label{fig:group_ambi_set}
    \vskip -0.2in
\end{figure}

Following~\cite{sagawa2019distributionally}, we define the unknown distribution $\tilde{\Pro}$ as a combination of $m$ distributions $\Pro_g$, each indexed by a group $g$ in the set $\mathcal{G} = \{1, 2, ..., m\}$. The ambiguity set $\mathfrak{M}$ is then defined as a convex combination of these groups:
\begin{equation}    \label{eq:group_ambi}
    \mathfrak{M} := \Big\{ \tilde{\Pro} = \sum_{g=1}^{m} q_g \, \Pro_g ~ | ~ q \in \Delta_m \Big\},
\end{equation}
where $\Delta_m$ denotes the $(m-1)$-dimensional probability simplex \cite{grunbaum1967convex} (see \cref{fig:group_ambi_set}).

In the dynamic chute mapping problem, we assume that past years of operational data from sortation warehouses provide sufficiently rich historical induction patterns that can be used to obtain representative distribution groups $\mathcal{G}$. With this assumption, it is reasonable to expect that any future induction pattern $\tilde{\Pro}$ can be represented as a combination of the basis distributions $\Pro_g$ with $g \in \mathcal{G}$. As shown in~\cite{sagawa2019distributionally}, evaluating the worst-case reward all $m$ groups in $\mathcal{G}$ is equivalent to evaluating the reward for the worst-case distribution within the ambiguity set $\mathfrak{M}$ defined in~\eqref{eq:group_ambi}.

\begin{lemma} \label{lem:group_dro_reward}
    Consider an ambiguity set $\mathfrak{M}$ formed by $\Pro_g$s as defined in~\eqref{eq:group_ambi}. For any state-action pair $(s,a) \in \mathcal{S}\times \mathcal{A}$, the worst-case expected reward satisfies:
    \begin{equation} \label{eq:lp_group_reduction}
         \inf_{g \in \mathcal{G}} \E_{X \sim \Pro_g} \left[ r\big(s,a;X \big)\right] = \inf_{\Pro \in \mathfrak{M}} \E_{X \sim \Pro} \left[ r\big(s,a;X \big)\right],
    \end{equation}
    where $\mathcal{G}$ denotes the set of group indices.
\end{lemma}

\Cref{lem:group_dro_reward} demonstrates a key advantage of group DRO: while general DRO problems are infinite-dimensional and computationally challenging, group DRO reduces the optimization to a finite-dimensional problem over $m$ groups. This reduction makes the training of distributionally robust MARL (DRMARL) computationally tractable.

\subsection{DRMARL with Group DRO}

In the MARL framework described in \cref{sec:problem_formulation}, the policy parameters $\theta$ are optimized as follows:
\begin{equation}
    \theta^* := \underset{\theta \in \Theta}{\arg \min} ~ \E_{X \sim \Pro} \left[ L(\theta;X) \right],
\end{equation}
and directly applying group DRO to MARL leads to:
\begin{equation} \label{eq:regular_group_DRO_loss}
    \tilde{\theta} := \underset{\theta \in \Theta}{\arg \min} \left\{ \underset{g \in \mathcal{G}}{\max} \, \E_{X \sim \Pro_g} \left[ L(\theta;X) \right] \right\}.
\end{equation}

However, conventional group DRO approaches are not directly applicable to MARL problems, since minimizing the worst-case Bellman error does not necessarily lead to a policy that is optimal under worst-case rewards. Instead, our proposed DRMARL seeks a robust policy that selects actions that are optimal with respect to worst-case reward functions. To achieve this, we introduce the distributionally robust Bellman operator, defined in the following result.

\begin{lemma} \label{lem:dr_bellman_operator}
    For an ambiguity set $\mathfrak{M}$ defined in~\eqref{eq:group_ambi} with group set $\mathcal{G}$, the distributionally robust Bellman operator is given by:
    \begin{equation*} \label{eq:dr_bellman_operator}
        \tilde{\mathcal{T}}_{\mathcal{G}} (\tilde{Q})(s, a) = \inf_{g \in \mathcal{G}} \E_{X \sim \Pro_g} [r(s, a;X)] + \gamma \max_{a'} \tilde{Q}(s', a'),
    \end{equation*}
    where $\tilde{Q}$ is the distributionally robust Q-function.
\end{lemma}

Accordingly, the distributionally robust loss is given by
\begin{equation} \label{eq:dr_loss}
    \begin{aligned}
        \tilde{L}(\theta;X) := \E_{s,a,r,a'} \big[ \big( \tilde{Q}(s,a;\theta) - \inf_{g \in \mathcal{G}} \E_{X \sim \Pro_g} [r(s, a;X)] - \gamma \, \underset{a'}{\max} \bar{\tilde{Q}} (s',a';\bar{\theta})\big)^2 \big],
    \end{aligned}
\end{equation}
and the distributionally robust parameters $\tilde{\theta}_{\mathcal{G}}$ are optimized by solving
\begin{equation}    \label{eq:regular_drmarl}
    \tilde{\theta}_{\mathcal{G}} := \underset{\theta \in \Theta}{\arg \min} ~ \tilde{L}(\theta;X).
\end{equation}


\section{Contextual Bandit-based Worst-Case Reward Predictor for DRMARL}    \label{sec:cb}

Solving the MARL group DRO problem \eqref{eq:regular_drmarl} is theoretically feasible since the worst-case reward can be evaluated for each $(s,a)$ by exhaustively searching among all distribution groups $\mathcal{G}$. However, this approach is inefficient when the number of groups is large and forward simulation in the environment is expensive. This is particularly the case in the multi-agent dynamic chute mapping problem, where millions of packages are sorted across many destinations. Common group DRO techniques, such as soft reweighting \cite{sagawa2019distributionally}, may not perform well in MARL because, unlike regression tasks, the data distribution depends on the agent's policy. As the policy evolves, the groups that are underrepresented or perform poorly may also change dynamically. This dynamic nature of the problem makes it challenging to apply static or even adaptive reweighting schemes, which assume a relatively stable data distribution.

To enhance the efficiency of the training process, we propose {\it a novel contextual bandit-based worst-case reward distribution predictor} that trains a contextual bandit (CB) \cite{li2010contextual,russo2018tutorial} model to predict the worst-case distribution group $g \in \mathcal{G}$ for each state-action pair $(s,a)$.

\begin{algorithm}[t]
\caption{CB-based Worst-Case Reward Predictor}
\label{alg:CB}
\begin{algorithmic}[1]
\STATE \textbf{Input:} Learning rate $l_{\textup{CB}}$, initial parameters $\psi_0$, induction distribution groups $\mathcal{G}$, MARL policy with $Q_{\textup{MARL}}$, exploration rate $\varepsilon_{\textup{CB}}$
\STATE \textbf{Initialize:} $\psi \gets \psi_0$, replay buffer $\mathcal{D}_{\textup{CB}} \gets \emptyset$
\FOR{episode $= 1,...,k_{\textup{CB}}$}
    \STATE Initialize the environment with random group $g' \in \mathcal{G}$ and observe initial state $s_0$
    \FOR{time step $t = 0,...,T$}
    \STATE Select action $a_t \gets \arg\max_{a \in \mathcal{A}} Q_{\textup{MARL}}(s_t,a)$ 
    \STATE With probability $\varepsilon_{\textup{CB}}$, select $g' \sim \text{Uniform}(\mathcal{G})$; otherwise, $g' \gets \arg\min_{g \in \mathcal{G}} \qcb(s_t,a_t,g;\psi)$
    \STATE Execute action $a_t$, observe reward $r_t$ and next state $s_{t+1}$ using group $g'$
    \STATE Store transition $(s_t, a_t, g', r_t, s_{t+1})$ in $\mathcal{D}_\textup{CB}$
    \STATE Sample mini-batch from $\mathcal{D}_{\textup{CB}}$ and update $\psi$:
    \STATE $\psi \gets \psi - l_{\textup{CB}} \nabla_\psi L_{\textup{CB}}(\psi)$
    \STATE $s_t \gets s_{t+1}$, reduce $\varepsilon_{\textup{CB}}$
    \ENDFOR
\ENDFOR
\STATE \textbf{Output:} Optimized CB predictor parameters $\psi^*$
\end{algorithmic}
\end{algorithm}

\subsection{CB-based Worst-Case Reward Predictor}

The CB treats the current state-action pair $(s,a)$ as contextual information, and its arms are defined over the groups of distributions in the set $\mathcal{G}$. The CB aims to predict the group $g$ that minimizes the reward $\E_{X \sim \Pro_{g}}r(s ,a ;X)$, which represents the worst-case reward among all groups. 
The CB is constructed as follows:

\textbf{Context space:} $\mathcal{S} \times \mathcal{A}$, where $(s,a) \in \mathcal{S} \times \mathcal{A}$ represents the state and the chute mapping actions at each step.

\textbf{Action space:} $\mathcal{G} = \{1, 2, \dots, m\}$, where $g \in \mathcal{G}$ denotes a group associated with an induction distribution.

\textbf{Reward:} A reward function $r: (\mathcal{S} \times \mathcal{A}) \times \mathcal{G} \to \mathbb{R}$, where $r(s ,a ;X)$ represents the observed reward for choosing distribution $\Pro_g$ at the current state-action pair $(s ,a)$.

The CB is represented by a $Q$-function that approximates the expected reward of choosing distribution $\Pro_g$ given a context $(s ,a)$. For this purpose, we use an independent DQN \cite{mnih2015human}:
\begin{equation}
    \qcb(s ,a ,g;\psi) = \E_{X\sim \Pro_g} \left[r(s ,a ;X) \right],
\end{equation}
where the reward function $r(s ,a ;X)$ is observed after running a single-step forward simulation with $(s ,a)$ under the induction-generating distribution $\Pro_g$. The $\qcb$ function is learned by minimizing the following loss, with the detailed training process shown in Algorithm~\ref{alg:CB}:
\begin{equation}    \label{eq:cb_loss}
    \begin{aligned}
    L_{\textup{CB}}(\psi) := ~ \E_{s,a} \big[ \big( \qcb \big(s,a,g;\psi \big) -  \E_{X\sim \Pro_g} \left[r(s ,a ;X) \right] \big)^2\big].
    \end{aligned}
\end{equation}

In Algorithm \ref{alg:CB}, the exploration of state-action pairs $(s,a)$ is guided by the existing MARL policy with $Q_{\textup{MARL}}$, which ensures sufficient exploration of the context space for the chute-mapping problem. For other applications, different exploration policies can be employed, such as random action selection, to ensure adequate coverage of the state-action space.

\subsection{DRMARL with CB-based Worst-Case Reward Predictor}
\begin{algorithm}[t]
\caption{DRMARL with CB-based Worst-Case Reward Predictor}
\label{alg:DRMARL}
\begin{algorithmic}[1]
\STATE \textbf{Input:} Learning rate $l_r$, initial parameters $\theta_0$, induction distribution groups $\mathcal{G}$, pre-trained CB predictor $\qcb$, exploration rate $\varepsilon$
\STATE \textbf{Initialize:} $\theta \gets \theta_0$, replay buffer $\mathcal{D} \gets \emptyset$
\FOR{episode $= 1,...,k$}
    \STATE Initialize the environment with random group $g' \in \mathcal{G}$ and observe initial state $s_0$
    \FOR{time step $t = 0,...,T$}
    \STATE With probability $\varepsilon$, select $a_t \sim \text{Uniform}(\mathcal{A})$; otherwise, $a_t \gets \arg\max_{a \in \mathcal{A}} \tilde{Q}(s_t,a;\theta)$
    \STATE Predict worst-case distribution group using CB: $g' \gets \arg\min_{g \in \mathcal{G}} \qcb(s_t,a_t,g)$
    \STATE Execute $a_t$, observe reward $r_t$ and next state $s_{t+1}$ using group $g'$
    \STATE Store transition $(s_t, a_t, g', r_t, s_{t+1})$ in $\mathcal{D}$
    \STATE Sample mini-batch from $\mathcal{D}$ and update parameters: $\theta \gets \theta - l_r \nabla_\theta \tilde{L}(\theta)$
    \STATE $s_t \gets s_{t+1}$, reduce $\varepsilon$
    \ENDFOR
\ENDFOR
\STATE \textbf{Output:} Optimized DRMARL policy parameters $\tilde{\theta}_{\mathcal{G}}$
\end{algorithmic}
\end{algorithm}

Once the $\qcb$ function has been trained, we can rewrite the distributionally robust loss \eqref{eq:dr_loss} as:
\begin{equation}    \label{eq:drcb_loss}
    \begin{aligned}
        \tilde{L}(\theta;X) = \E_{s,a,r,a'} \Big[ \big( \tilde{Q}(s,a;\theta) -  \E_{X \sim \Pro_{g'}} \left[ r(s,a;X) \right] - \gamma \, \underset{a'}{\max} \bar{\tilde{Q}} (s',a';\bar{\theta})\big)^2 \Big],
    \end{aligned}
\end{equation}
where $g' = \arg\min_{g \in \mathcal{G}} \qcb(s,a,g)$ is the index of the distribution group with the predicted worst-case reward. 

The training procedure of DRMARL is shown in Algorithm~\ref{alg:DRMARL}. The key difference compared to MARL training is that DRMARL aims to train a DQN that estimates the worst-case return among all groups, while MARL aims to estimate the observed return only for a specific induction distribution. Moreover, in contrast to traditional group DRO, the index of the worst-case distribution $g'$ in DRMARL is not obtained via exhaustive search; instead, it is predicted by $\qcb$ given the state-action pair $(s,a)$. The independent $Q$-network $\qcb$ is learned beforehand using Algorithm \ref{alg:CB} and remains unchanged during DRMARL training. While it may seem counter-intuitive that $\qcb$ predicts the worst-case group after $\tilde{Q}$ selects an action, this design enables $\qcb$ to provide the worst-case expected return for each $(s,a)$ pair, thereby enabling the learning of a robust policy.


\section{Simulation}    \label{sec:simulation}

In this section, we demonstrate the effectiveness of the proposed DRMARL policy under adversarial induction changes in both a simplified simulation and a large-scale Amazon robotic sortation warehouse simulation environment.

\subsection{Simplified Robotic Sortation Warehouse
Simulation}   \label{sec:simple_simulation}

\begin{table}[t]
\caption{Key sortation metrics comparison across policies, averaged over $m=9$ groups.}
\label{tab:simple_end_metric}
\centering
\resizebox{0.8 \linewidth}{!}{%
\begin{tabular}{lccc}
\toprule
\textbf{Policy} & \textbf{\makecell{Recirculation  Rate ($\downarrow$)}} & \textbf{Throughput ($\uparrow$)}  & \textbf{\makecell{ Recirculation  Amount ($\downarrow$)}}  \\ 
\midrule
MARL & $2.16\% \pm 2.35 \%$ & $11740.98 $ & $259.02 $ \\ 
\makecell[l]{DRMARL (random)}  & $1.56\% \pm 1.45\%$ & $11812.77$ & $187.23$ \\ 
\makecell[l]{DRMARL (with $\qcb$)}  & $\bm{0.56\% \pm 0.18\%}$ & $\bm{11932.21}$ & $\bm{67.79}$ \\ \hline
\makecell[l]{DRMARL (exhaustive)}  & $0.55\% \pm 0.13\%$ & $11933.68$ & $66.32$ \\
\makecell[l]{MARL (group-specific)} & $0.53\% \pm 0.14 \%$ & $11936.60 $ & $63.40 $ \\
\bottomrule
\end{tabular}%
}
\end{table}

In the simplified simulation environment, there are $10$ eject chutes, one recirculation chute, and $20$ total unique destinations. Packages arrive at the sortation warehouse according to the induction data $X$ generated from an induction distribution $\Pro$. When packages exceed the chutes' capacities, they are sent to the recirculation chute. One training/testing episode consists of $5$ hours, with each time step being $30$ minutes, after which the environment is reset. An eject chute can be reallocated at each time step. The implementation details are provided in \cref{app:simple_structure}.

We train the DRMARL policy over 300 episodes using training data generated from $9$ distinct induction distribution groups. Similarly, the regular MARL policy is trained for 300 episodes on the same groups. Due to the stochastic nature of the induction-generating distributions, the induction data differs for each simulation trial.

The comparison of key metrics is shown in \cref{tab:simple_end_metric}. The DRMARL policy with $\qcb$ achieves the best performance across all metrics. For reference, the last row shows the theoretical optimal performance of group-specific MARL policies, which are trained and tested on the same group. DRMARL performs only {\it marginally} below this optimal baseline, demonstrating a favorable trade-off between individual performance and distributional robustness. Comparison with both random group selection and exhaustive worst-case search demonstrates that $\qcb$ efficiently identifies worst-case groups while learning an equally effective policy, performing significantly better than random selection and matching the robustness of exhaustive search. Here, random group selection refers to replacing Line 7 of \cref{alg:DRMARL} with:
\[
g' \gets \text{Uniform}(\mathcal{G})
\]
instead of using:
\[
g' \gets \arg\min_{g \in \mathcal{G}} \qcb(s_t,a_t,g)
\]
This confirms that $\qcb$ effectively enables the DRMARL policy to explore worst-case reward functions during training. Furthermore, when compared to a DRMARL policy trained using exhaustive search over worst-case rewards for each state-action pair $(s,a)$, the $\qcb$-based approach achieves equivalent policy performance and robustness while being computationally more efficient. The last row presents the theoretical optimal performance of group-specific MARL policies (trained and tested on the same group). DRMARL performs only marginally below this optimal baseline, demonstrating an effective balance between individual performance and distributional robustness.

\begin{figure}[t]
    \vskip 0.2in
    \centering
    \includegraphics[width = 0.7 \linewidth]{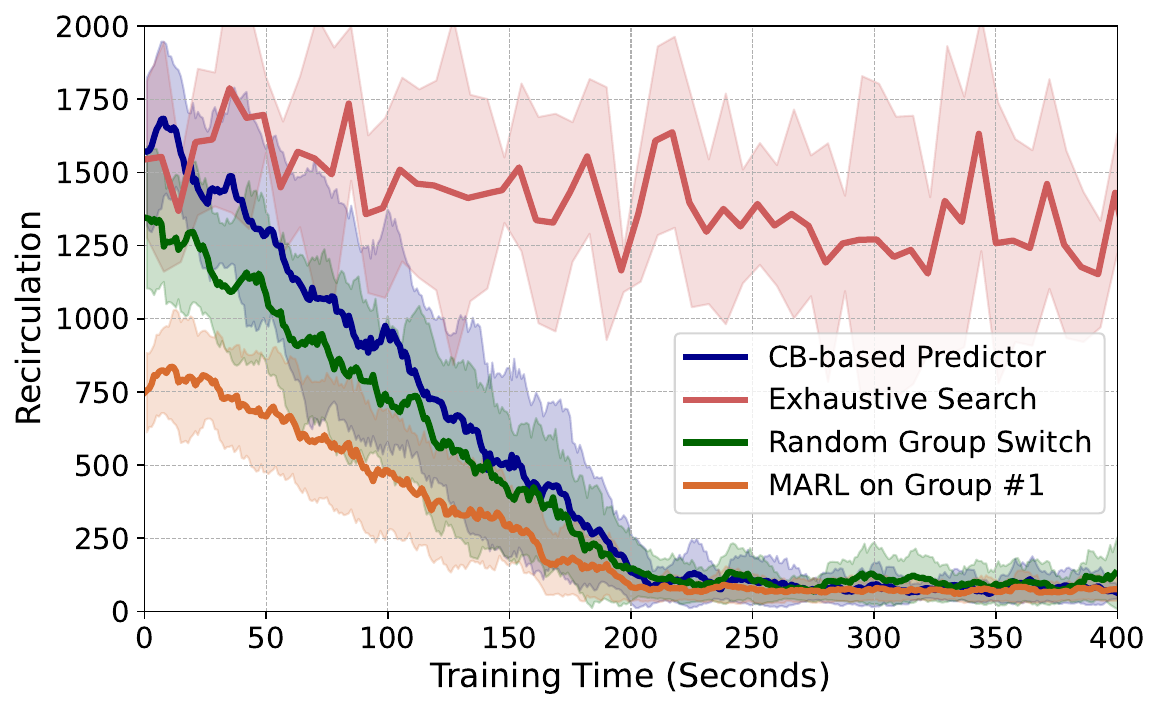}
    \caption{Training efficiency comparison in simplified warehouse.}
    \label{fig:simple_cb_efficiency}
    \vskip -0.2in
\end{figure}
\begin{figure}[t]
    \vskip 0.2in
    \centering
    \includegraphics[width=0.7\linewidth]{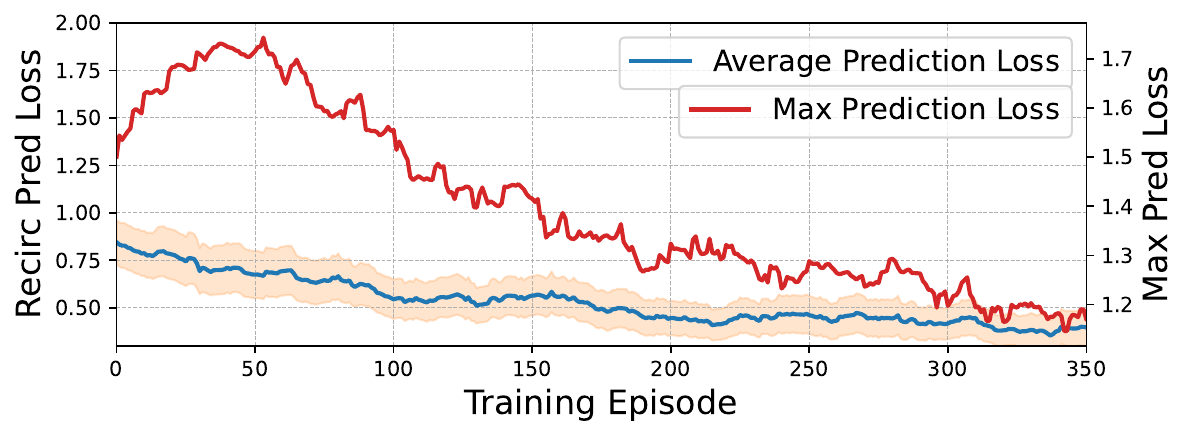}
    \caption{Training recirculation rate prediction loss \eqref{eq:cb_loss} for CB in simplified robotic sortation warehouse.}
    \label{fig:simple_cb_train}
    \vskip -0.2in
\end{figure}

\cref{fig:simple_cb_train} illustrates the training progress of $\qcb$, showing significant reductions in both average and maximum prediction errors of the recirculation rate throughout the training process. The training efficiency comparison across different approaches is presented in \cref{fig:simple_cb_efficiency}. DRMARL with $\qcb$ requires less than $300$ seconds to converge, while an exhaustive search for the worst-case group takes at least $2900$ seconds to complete. DRMARL with exhaustive search requires the longest training time due to its comprehensive exploration across all distribution groups. In contrast, DRMARL with $\qcb$ converges significantly faster while matching the exhaustive search's recirculation performance in initial stages of the training, validating $\qcb$'s ability to identify worst-case groups. DRMARL with random group selection shows faster convergence than the $\qcb$-based approach, but this is because random exploration does not guarantee finding worst-case reward functions. The group-specific MARL policy exhibits the fastest convergence due to the relative simplicity of its training task.

\subsection{Large-Scale Amazon Robotic Sortation Warehouse Simulation}
\subsubsection{Implementation Details}

\begin{figure}[t]
    \centering
    \includegraphics[width = 0.5 \linewidth]{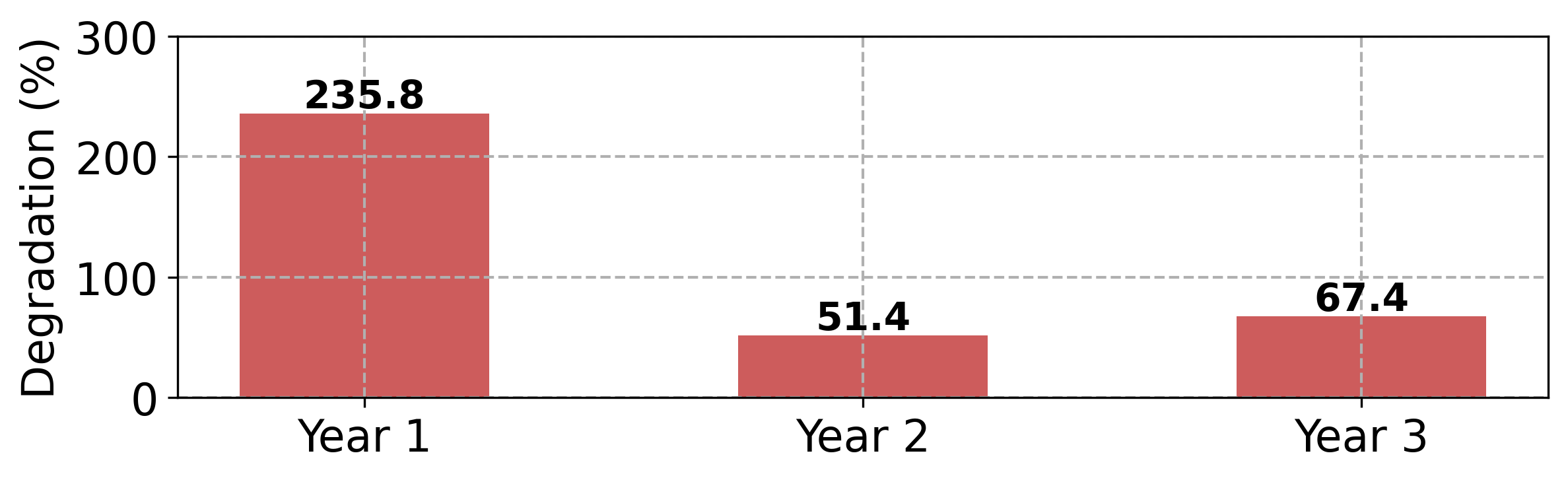}
    \caption{Relative degradation of MARL policy (trained on Year 4) on OOD (Years 1-3) induction data, compared to Year 4.}
    \label{fig:OOD_degradation}
\end{figure}

In the large-scale simulation environment, there are $187$ eject chutes, one recirculation chute, and $120$ total unique destinations. Packages arrive at the sortation warehouse according to the corresponding induction data $X$ generated from the induction distribution $\Pro$\footnote{Due to common industry confidentiality practices, we cannot disclose the specific data source and report only relative performance improvements. The data represents realistic package flow patterns typical of Amazon robotic sortation facilities.}. When packages exceed chute capacities or miss departure transportation schedules, they are sent to the recirculation chute and added to the sequence of new packages at the next time step. One training/testing episode consists of $11$ hours, with each time step being five minutes, after which the environment is reset. Every five minutes, we assign destinations to chutes that become available for reallocation.

We train the DRMARL policy over 200 episodes using training data generated from 21 distinct induction distribution groups spanning several years. Similarly, the regular MARL policy is trained for 200 episodes using induction data from Year 4. For testing, we evaluate both policies on newly generated induction data from 21 distinct distribution groups, conducting five experiments per generated induction. Due to the stochastic nature of the induction-generating distributions, the test induction data remains unseen during training for both policies.

\subsubsection{Robustness of the Chute Mapping Policy}

\begin{table}[t]
\caption{Relative improvement ($\uparrow$) of DRMARL over MARL baseline, averaged across $m = 21$ groups.}
\label{tab:end_metric}
\centering
\resizebox{0.8 \linewidth}{!}{%
\begin{tabular}{lccc}
\toprule
\textbf{Policy} & \textbf{\makecell{Recirc Rate Reduction}} & \textbf{\makecell{Throughput Increase}}  & \textbf{\makecell{ Recirc Amount Reduction}}  \\ 
\midrule
DRMARL  & $79.97\%$ & $5.62 \%$ & $33.64 \%$ \\ \hline
\makecell[l]{MARL (group-specific)} & $85.42 \%$ & $9.80 \%$ & $40.50 \%$ \\ 
\bottomrule
\end{tabular}%
}
\end{table}

To motivate the need for the proposed DRMARL framework, we first study the performance of the regular MARL policy on out-of-distribution (OOD) induction data (e.g., induction data from Years 1-3), as shown in \cref{fig:OOD_degradation}. The results suggest that the regular MARL policy is not robust against OOD induction data.

\Cref{tab:end_metric} presents the average relative performance improvement of DRMARL across all $21$ distribution groups, using MARL as the baseline. The proposed DRMARL method demonstrates robust performance on all induction groups, consistently outperforming the baseline MARL policy. For reference, the bottom row shows the theoretical optimal performance obtained by training and testing a group-specific MARL on each individual group. As expected, DRMARL performs {\it marginally} below these group-specific MARL policies, illustrating the trade-off between performance and distributional robustness. Detailed results are provided in Appendix~\ref{app:detail_validation}.

\begin{figure}[t]
    \centering
    \includegraphics[width = 0.6 \linewidth]{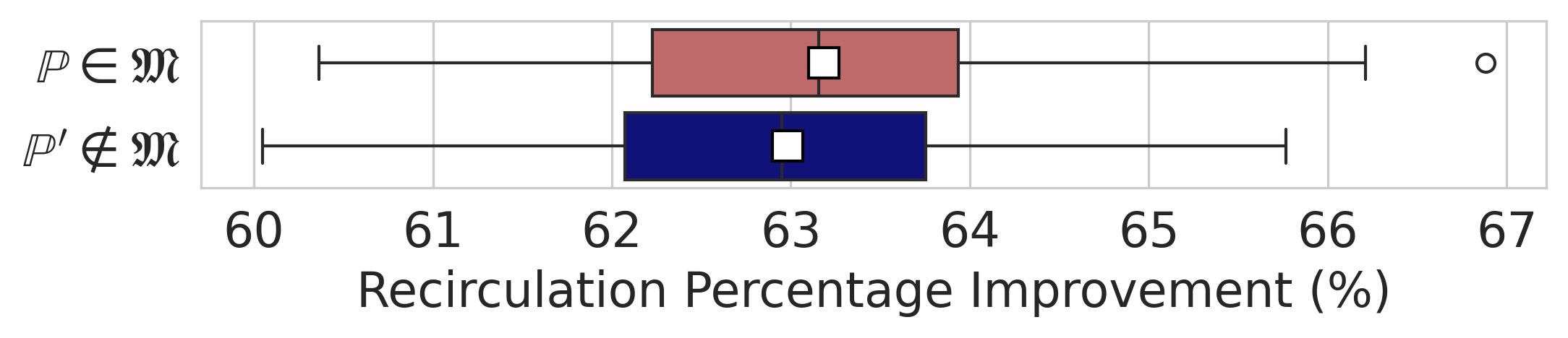}
    \caption{Recirculation rate improvement of DRMARL over two equally-performing MARL policies trained on distributions inside and outside $\mathfrak{M}$.}
    \label{fig:DR_outside_M}
\end{figure}

To evaluate DRMARL's robustness beyond the ambiguity set $\mathfrak{M}$, we tested it on induction distributions $\Pro'$ outside the group ambiguity set $\mathfrak{M}$ ($\Pro' \notin \mathfrak{M}$). As shown in \cref{fig:DR_outside_M}, DRMARL maintains consistent sortation performance and generalizes effectively to distributions even outside the ambiguity set $\mathfrak{M}$.

\subsubsection{CB-based Worst-Case Reward Predictor}

\begin{figure}[t]
    \centering
    \includegraphics[width = 0.7 \linewidth]{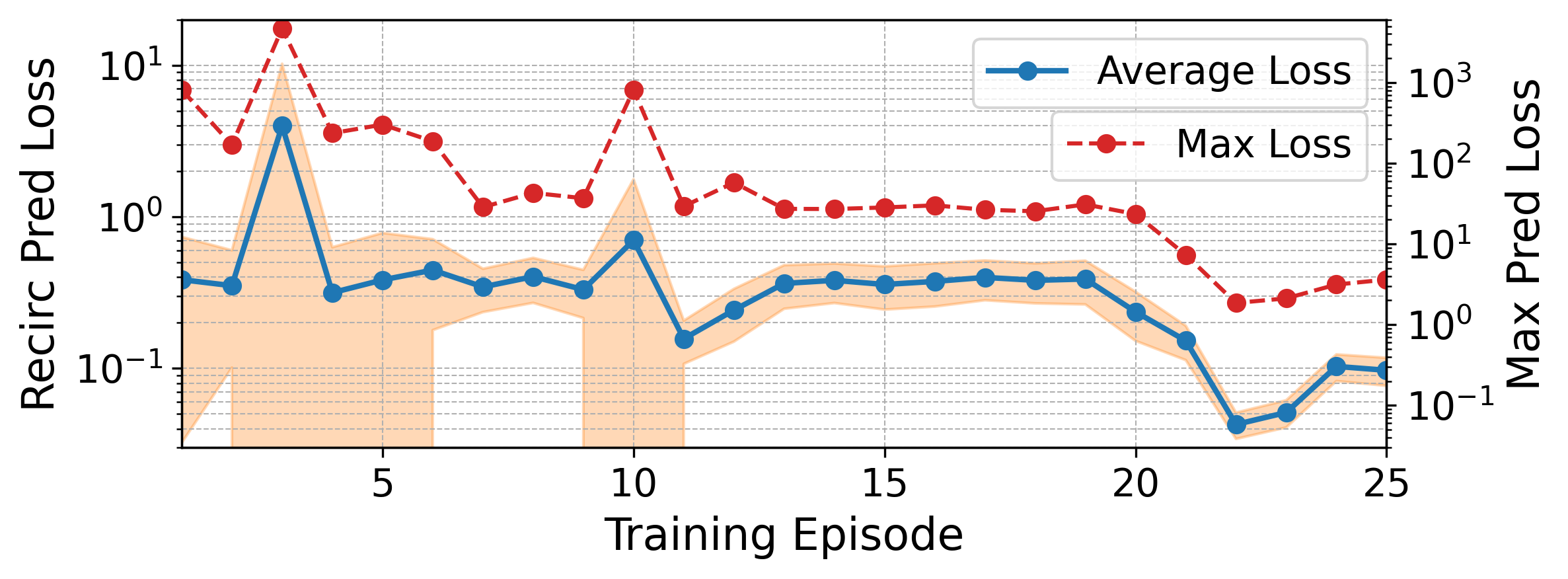}
    \caption{Training prediction loss \eqref{eq:cb_loss} of $\qcb$ predictor, averaged over 11-hour simulations.}
    \label{fig:cb_train}
\end{figure}

\begin{figure}[t]
    \centering
    \includegraphics[width = 0.7 \linewidth]{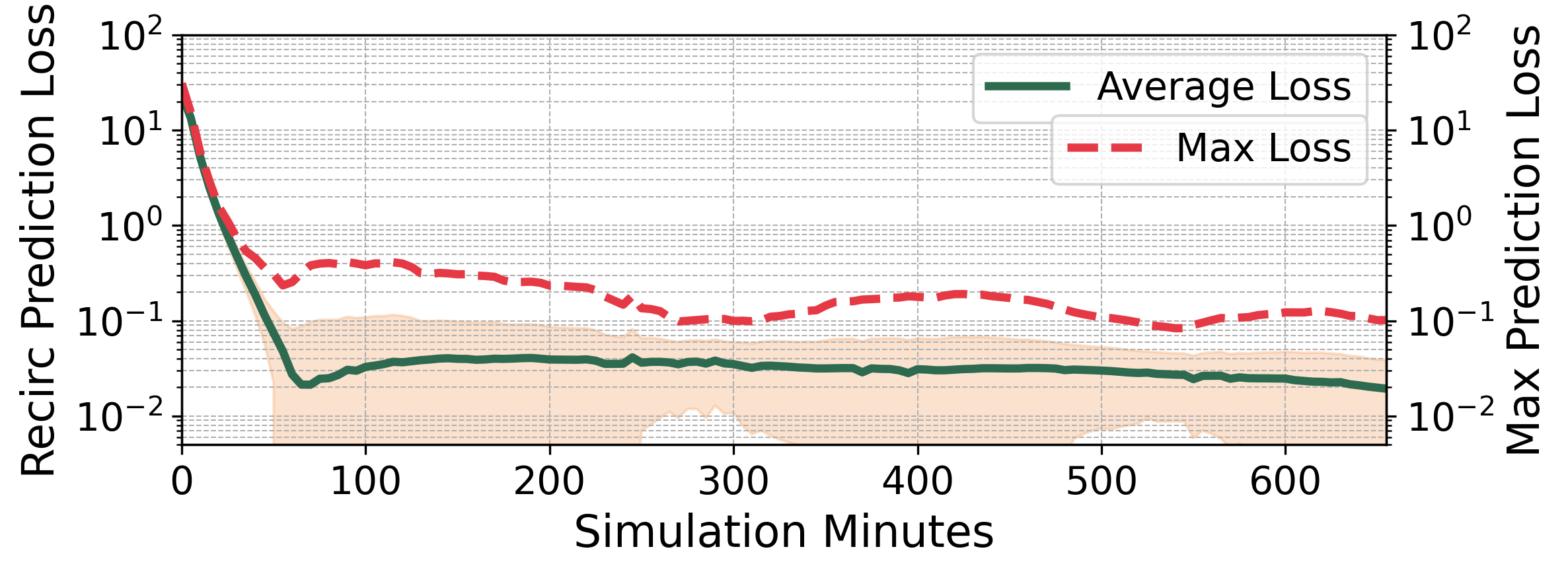}
    \caption{$\qcb$ prediction loss for recirculation percentage, averaged over 25 test simulations.}
    \label{fig:cb_pred_loss}
\end{figure}

Following \cref{sec:cb}, we train an independent Q-network $\qcb$ to predict the worst-case recirculation (reward) for each state-action pair $(s,a)$ across groups $\mathcal{G}$. The training of $\qcb$ uses the existing MARL policy, with the progress shown in \cref{fig:cb_train}. The learned $\qcb$ achieves high accuracy, with prediction errors below $1\%$ of the recirculation rate, enabling reliable identification of worst-case scenarios among groups $g \in \mathcal{G}$.

During each day's $11$-hour simulation, as illustrated in \cref{fig:cb_pred_loss}, $\qcb$'s prediction accuracy improves substantially after the first hour. While initially suboptimal, $\qcb$'s performance remains sufficient for DRMARL training, as the impact of worst-case distributions on recirculation becomes more pronounced in later stages when fewer chutes are available.

\subsubsection{Efficient Training with CB-based Worst-Case Reward Predictor}
\begin{figure}[t]
    \vskip 0.2in
    \centering
    \includegraphics[width = 0.7 \linewidth]{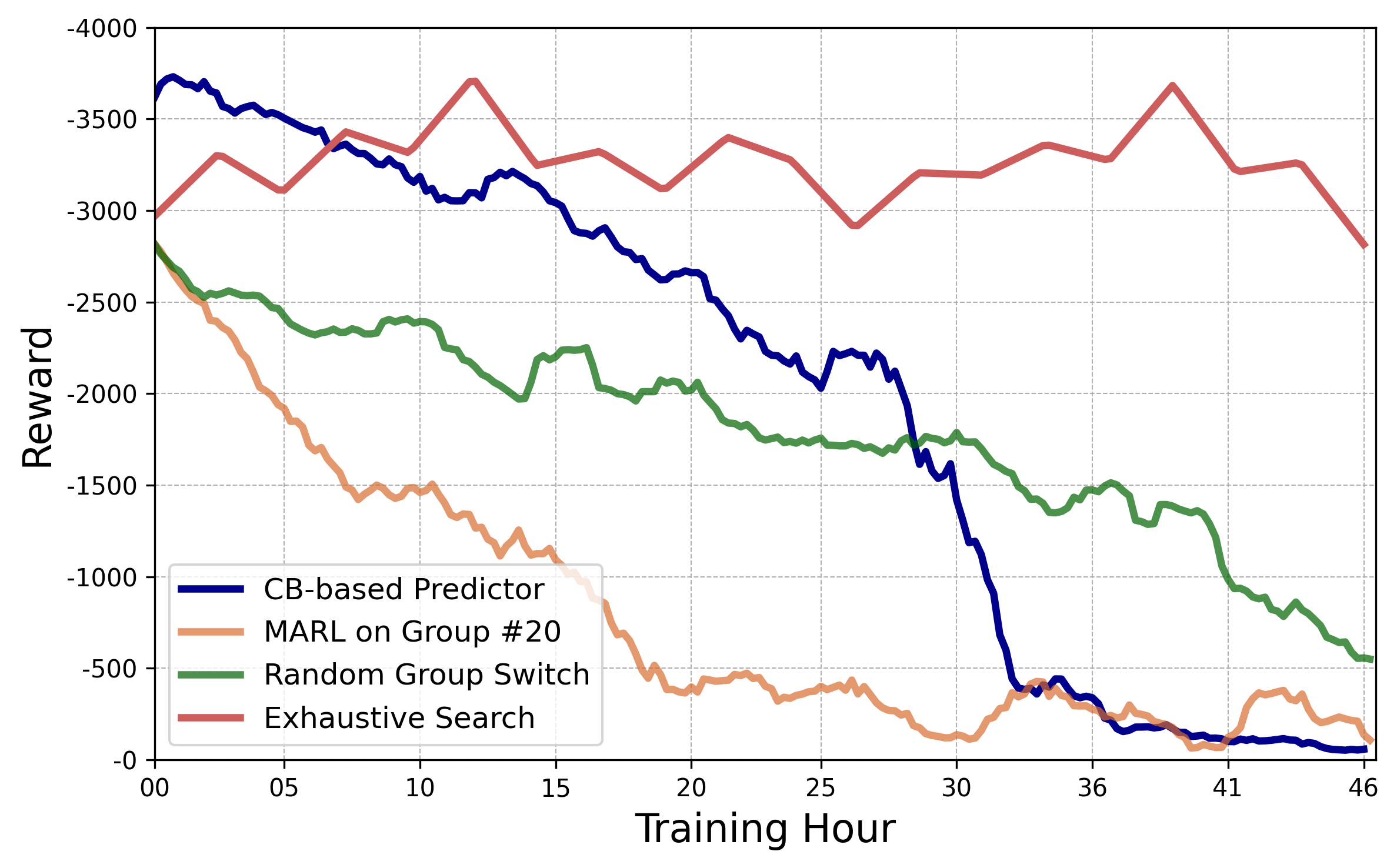}
    \caption{Training computational efficiency comparison in large-scale Amazon robotic sortation warehouse simulation environments.}
    \label{fig:CB_efficiency}
    \vskip -0.2in
\end{figure}

The CB-based worst-case reward predictor $\qcb$ substantially improves training efficiency by eliminating the need for exhaustive group evaluation at each time step, reducing the computational complexity of the worst-case group identification from $\mathcal{O}(m)$ to $\mathcal{O}(1)$. As demonstrated in \cref{fig:CB_efficiency}, training with $\qcb$ achieves significantly faster convergence compared to exhaustive evaluation over $\mathcal{G}$, which requires approximately $924$ hours on a cloud instance with 64 vCPUs (Intel Xeon Scalable 4th generation) and 128 GB RAM. The lightweight $Q$-network updates enabled CPU-only training, with most computation time spent on environment simulation. This efficiency advantage becomes even more pronounced when dealing with complex environments or larger group sets.

\cref{fig:CB_efficiency} also compares the training efficiency of different approaches. The group-specific MARL, which trains on group $\# 20$, shows the fastest convergence due to its simplified learning objective. Among distributionally robust approaches, DRMARL with random group selection ($g' \gets \text{random}(\mathcal{G})$) converges initially faster than other variants but achieves suboptimal robustness since it may miss critical worst-case scenarios. DRMARL with $\qcb$ strikes a balance between training speed and performance, converging significantly faster than exhaustive search while maintaining near-optimal worst-case performance guarantees. As expected, DRMARL with exhaustive search over all distribution groups requires the longest training time, though it serves as a valuable baseline for validating the efficiency of our $\qcb$-based approach.


\section{Conclusion} \label{sec:conclusion.}
In this paper, we introduced DRMARL, a framework that integrates group-DRO into MARL to enhance policy robustness against adversarial distribution shifts in warehouse sortation systems. To address the computational cost of identifying the worst-case group, we developed a CB-based predictor that significantly reduces worst-case identification complexity from $\mathcal{O}(m)$ to $\mathcal{O}(1)$. Experimental results from both simplified and large-scale warehouse environments demonstrate that DRMARL achieves near-optimal performance across all distribution groups while maintaining computational efficiency. The framework shows strong generalization even to distributions outside the training set, and its design principles can be extended to other MARL applications where distributional robustness is crucial.

\section*{Acknowledgments} \label{sec:acknow}
We would like to express our sincere gratitude to our colleagues at Amazon for their support and valuable contributions throughout this research. In particular, we extend special thanks to Rahul Chandan and Mouhacine Benosman for their insightful feedback and constructive discussions to this work.

\printbibliography

\newpage
\appendix
\onecolumn

\section*{Appendix}

\section{Proofs of Theoretical Results}     \label{app:proofs}
\subsection{Proof of \cref{lem:group_dro_reward}}
Recall the definition of $\mathfrak{M}$ in \eqref{eq:group_ambi} and for any probability distribution $\Pro \in \mathfrak{M}$, we have 
\begin{equation}
    \begin{aligned}
        \E_{X \sim \Pro} \left[ r\big(s,a;X \big)\right] &= \int  r\big(s,a;X \big) \, \Pro(X) \, d X\\
        & = \int  r\big(s,a;X \big) \, \left(\sum_{g = 1}^{m} q_g \, \Pro_i(X) \right) \, d X\\
        & = \sum_{g = 1}^{m} q_g  \int  r\big(s,a;X \big) \, \Pro_i(X) \, d X = \sum_{g = 1}^{m} q_g \, \E_{X \sim \Pro_g} \left[ r\big(s,a;X \big)\right].
    \end{aligned}
\end{equation}
Then, taking the infimum on both sides yields:
\begin{equation}
    \begin{aligned}
        \inf_{\Pro \in \mathfrak{M}} \E_{X \sim \Pro} \left[ r\big(s,a;X \big)\right] &= \inf_{q \in \Delta_m} \sum_{g = 1}^{m} q_g \E_{X \sim \Pro_g} \left[ r\big(s,a;X \big)\right] \\
        &= \inf_{g \in \mathcal{G}} \E_{X \sim \Pro_g} \left[ r\big(s,a;X \big)\right],
    \end{aligned}
\end{equation}
since optimum of a linear program over simplex $\Delta_m$ is obtained at vertices. \hfill$\blacksquare$

\subsection{Proof of \cref{lem:dr_bellman_operator}}
In order to find the group distributionally robust action-value function $\tilde{Q}$, we consider the worst-case immediate reward at each state-action pair $(s,a)$ as:
\begin{equation}
    \tilde{r}(s,a) = \min_{g \in \mathcal{G}} \E_{X \sim \Pro_g} \left[ r(s,a;X) \right] \leq \E_{X \sim \Pro} \left[ r(s,a;X) \right],
\end{equation}
for all unknown distribution $\Pro \in \mathfrak{M}$. Then, the worst-case return is given by: 
\begin{equation}
    \tilde{R}_{t} = \sum_{k = 0}^{\infty} \gamma^{k} \, \tilde{r}(s,a) \leq \sum_{k = 0}^{\infty} \gamma^{k} \, \E_{X \sim \Pro_k}[r_{t+k+1}(s,a;X)],
\end{equation}
where $\gamma \in [0,1]$ is the discount factor and the inequality holds for all possible sequences $\{\Pro_k\}_{k=0}^{\infty}$ with $\Pro_k \in \mathfrak{M}$. Then, the worst-case action-value function under policy $\pi$ can be expressed as:
\begin{equation}
    \begin{aligned}
        \tilde{Q}^{\pi}(s,a) &= \E_{\pi} \left[ \tilde{r}(s,a) + \gamma \tilde{Q}^{\pi}(s_{t+1},a_{t+1}) ~\big|~ s_t = s, a_t = a \right]\\
        &= \E_{s' \sim P(\cdot|s,a)} \left[ \tilde{r}(s,a) + \gamma \E_{a' \sim \pi(\cdot|s')} \tilde{Q}^{\pi}(s',a') \right],
    \end{aligned}
\end{equation}
and the optimal worst-case action-value function satisfies the Bellman optimality equation:
\begin{equation}
    \tilde{Q}^*(s,a) = \tilde{r}(s,a) + \gamma \, \E_{s' \sim P(\cdot|s,a)} \left[ \max_{a'} \tilde{Q}^*(s',a') \right].
\end{equation}

Following the result from \cite{liu2022distributionally}, the distributionally robust Bellman Operator with ambiguity sets $\mathfrak{P}$ and $\mathfrak{R}$ is given by 
\begin{equation}
    \tilde{\mathcal{T}}_{\mathfrak{P},\mathfrak{R}}( \tilde{Q})(s, a) = \inf_{ \substack{ p_{s,a} \in \mathfrak{P} \\ r_{s,a} \in \mathfrak{R}}} \left\{ \E_{r_{s,a}} [r(s, a)] + \gamma \, \E_{p_{s,a}} \left[\max_{a'} \tilde{Q}(s', a') \right]\right\},
\end{equation}
where $p_{s,a}$ and $r_{s,a}$ denote the distributions of state transitions probabilities and reward functions respectively, with their corresponding ambiguity sets $\mathfrak{P}$ and $\mathfrak{R}$. Since the distribution shift in the random variable $X$ only affects the reward, i.e., the distribution of the reward function $r_{s,a}$, we have:
\begin{equation}
    \begin{aligned}
        \tilde{\mathcal{T}}_{\mathfrak{R}}(\tilde{Q})(s, a) &= \inf_{ r_{s,a} \in \mathfrak{R}} \left\{ \E_{r_{s,a}} [r(s, a)] + \gamma \, \left[\max_{a'} \tilde{Q}(s', a') \right]\right\}\\
        &=  \tilde{r}(s, a) + \gamma \, \max_{a'} \tilde{Q}(s', a') \\
        &= \inf_{ g \in \mathcal{G}} \left\{ \E_{X \sim \Pro_g} [r(s, a;X)]\right\} + \gamma \, \max_{a'} \tilde{Q}(s', a') =\tilde{\mathcal{T}}_{\mathcal{G}}(\tilde{Q})(s, a) .
    \end{aligned}
\end{equation}

We further show the distributionally robust Bellman operator is a contraction map under the $\ell_\infty$ norm. Consider two arbitrary robust action-value functions $\tilde{Q}_1$ and $\tilde{Q}_2$ such that 
\begin{equation}
    \begin{aligned}
        \tilde{\mathcal{T}}_{\mathcal{G}}(\tilde{Q}_1)(s, a) &= \tilde{r}(s, a) + \gamma \, \max_{a'} \tilde{Q}_1(s', a')\\
        \tilde{\mathcal{T}}_{\mathcal{G}}(\tilde{Q}_2)(s, a) &= \tilde{r}(s, a) + \gamma \, \max_{a'} \tilde{Q}_2(s', a').
    \end{aligned}
\end{equation}
Finding the difference yields
\begin{equation}
    |\tilde{\mathcal{T}}_{\mathcal{G}}(\tilde{Q}_1)(s, a) - \tilde{\mathcal{T}}_{\mathcal{G}}(\tilde{Q}_2)(s, a)| \leq \gamma \max_{s',a'} |\tilde{Q}_1(s', a') - \tilde{Q}_2(s', a')|,
\end{equation}
and taking the maximum over all feasible state-action pair $(s,a)$ implies
\begin{equation}
    \|\tilde{\mathcal{T}}_{\mathcal{G}}(\tilde{Q}_1) - \tilde{\mathcal{T}}_{\mathcal{G}}(\tilde{Q}_2)\|_{\ell_\infty} \leq \gamma \|\tilde{Q}_1 - \tilde{Q_2}\|_{\ell_\infty}.
\end{equation}
Since $\gamma \in [0,1]$, the robust Bellman operator is contraction map and the $Q$-learning algorithm will converge to $\tilde{Q}^*$. \hfill$\blacksquare$

\section{Implementation Details for Simplified Robotic Sortation Warehouse}
\label{app:simple_structure}

We define a Markov game for $N$ agents (representing unique destinations) by the tuple

\noindent$\left(N,\mathcal{S},\{\mathcal{O}^i\}_{i=1}^N,\{\mathcal{A}^i\}_{i=1}^N,P,\{r^i\}_{i=1}^N,\gamma,\rho_0 \right)$, where: 

\begin{itemize}
\item[(a)] \textbf{Agents:} The set of $N$ agents, each corresponding to a unique destination.

\item[(b)] \textbf{State Space:} $\mathcal{S}$ denotes the joint state space. 

\item[(c)] \textbf{Observation Space:} For each agent $i$, $\mathcal{O}^i\subset \mathcal{S}$ represents its local observation at each time step, consisting of:
\begin{itemize}
    \item The total number of assignable chutes (uniform across all agents)
    \item The number of chutes currently assigned to agent $i$
\end{itemize}

\item[(d)] \textbf{Action Space:} For each agent $i$, $\mathcal{A}^i\subset [0,1]$ represents its action space, where each action determines if a new chute will be allocated. An action value of 1 indicates the assignment of a new chute to destination $i$. The joint action space is defined as $\mathcal{A}=\prod_{i=1}^N \mathcal{A}^i$.

\item[(e)] \textbf{Transition Probability:} $P:\mathcal{S}\times \mathcal{A} \times \mathcal{S}\to [0,1]$ specifies the probability of transitioning between states, representing the likelihood of packages being either successfully sorted or diverted to the recirculation buffer.

\item[(f)] \textbf{Reward Function:} For each agent $i$, $r^i:\mathcal{S}\times \mathcal{A} \times \mathcal{X} \to\mathbb{R}$ defines the reward function, which penalizes the number of packages in the recirculation buffer resulting from the current chute allocation.
\end{itemize}

The model is completed with discount factor $\gamma\in(0,1)$ and initial state distribution $\rho_0$. In \cref{sec:simple_simulation}, we employ the Value Decomposition Network (VDN) \cite{shen2023multi} combined with budget constraints in computing joint actions. This approach addresses both the scalability challenges of the state-action space and the computational feasibility of the expectation in \eqref{eq:DQN_loss}. Detailed implementations are provided in~\cref{sec:vdn} and \cref{sec:mip}.

In the simplified robotic sortation environment, we fix the total induction volume at each time step to $1200$ packages. The number of incoming packages for each destination $i = 1,\ldots,N$ follows an unknown normal distribution $\mathcal{N}(\mu, \sigma)$. For each destination $i$, the probability that an incoming package is assigned to destination $i$ is given by:
\begin{equation}    \label{eq:simple_induct_sample}
    \Pro \Big\{ \textup{incoming package belongs to } i \Big\} = \frac{\Phi(\frac{i - \mu}{\sigma}) - \Phi(\frac{i-1 - \mu}{\sigma}) }{\Phi(\frac{N - \mu}{\sigma}) - \Phi(\frac{- \mu}{\sigma}) },
\end{equation}
where $\Phi(z) = \frac{1}{\sqrt{2\pi}} \int_{-\infty}^{z} e^{-\frac{t^2}{2}} \, dt$ is the cumulative distribution function of the standard normal distribution. This formulation ensures $\sum_{i=1}^{N} \Pro \{ \textup{incoming package belongs to } i \} = 1$. The distribution of packages across destinations is then determined by sampling $1200$ packages according to the probabilities defined in \eqref{eq:simple_induct_sample} at each time step.

In \cref{sec:simple_simulation}, we assume the destination transportation vehicle has infinite capacity, meaning packages enter the recirculation buffer only when incoming packages are destined for a location without an assigned eject chute. In this example, we construct the ambiguity set as:
\begin{equation}
    \mathfrak{M} := \Big\{ \tilde{\Pro} = \sum_{g=1}^{m} q_g \, \Pro_g ~ | ~ q \in \Delta_m \Big\},
\end{equation}
where each $\Pro_g$ represents a normal distribution $\mathcal{N}(\mu_g, \sigma)$. For our simulation, we construct the ambiguity set using $m = 9$ groups with means $\mu_g \in \{-4,-3,\ldots,0,\ldots,4\}$, standard deviation $\sigma = 2$, and index set $\mathcal{G} = \{1,2,\ldots,9\}$.

\section{Implementation Details for Large-Scale Amazon Robotic Sortation Warehouse}

\subsection{Distributionally Robust Bellman Operator in Large-Scale Robotic Sortation Warehouses}     \label{app:dr_bo_approximate}

\begin{figure}[t]
    \centering
    \includegraphics[width = \linewidth]{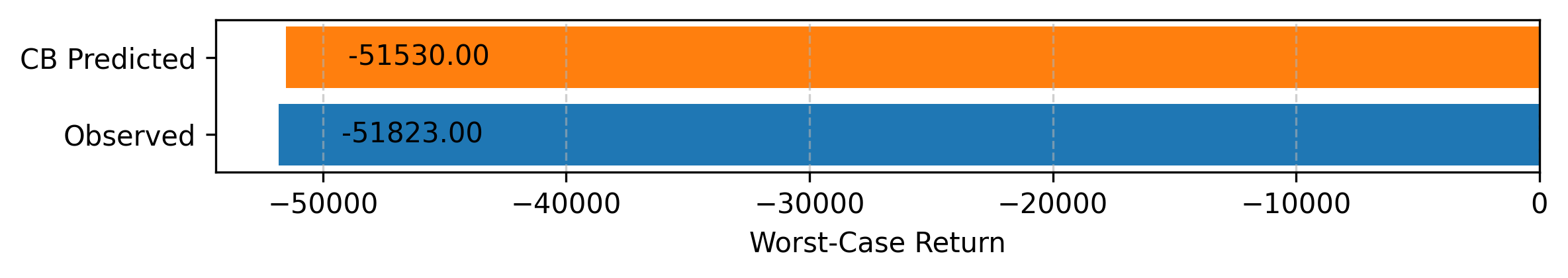}
    \caption{Comparison of worst-case returns: predictions using $\qcb$ for both immediate rewards and transition probabilities versus observed values from extensive state-action space exploration. The close alignment validates our approach of using $\qcb$ to approximate both components of the worst-case scenario \eqref{eq:dr_bo_approx}.}
    \label{fig:trans_prob_compare}
\end{figure}

In large-scale robotic sortation warehouses, we observe that the transition probability is dependent on the induction random variable $X$, which violates the assumption in Lemma~\ref{lem:dr_bellman_operator}. Consequently, we must compute:
\[
\inf_{\mathbb{P} \in \mathfrak{M}} \gamma \, \mathbb{E}_{p_{s,a}(X), X\sim \mathbb{P}_g} \left[\max_{a'} \tilde{Q}(s', a') \right]
\]
for the distributionally robust Bellman operator. This leads to:
\begin{equation}
    \label{eq:dr_bo_warehouse}
    \tilde{\mathcal{T}}_{\mathfrak{R}}(\tilde{Q})(s, a) = \inf_{g \in \mathcal{G}} \left\{ \mathbb{E}_{X \sim \mathbb{P}_g} [r(s, a;X)] \right\} + \inf_{\mathbb{P} \in \mathfrak{M}} \left\{\gamma \, \mathbb{E}_{p_{s,a}(X), X\sim \mathbb{P}_g}\left[\max_{a'} \tilde{Q}(s', a')  \right] \right\}
\end{equation}
where the computation becomes infinite-dimensional and practically intractable. To address this, during training, we approximate the distributionally robust Bellman operator with:
\begin{equation}    
    \label{eq:dr_bo_approx}
    \tilde{\mathcal{U}}_{\mathfrak{R}}(\tilde{Q})(s, a) = \inf_{g \in \mathcal{G}} \left\{ \mathbb{E}_{X \sim \mathbb{P}_g} [r(s, a;X)] + \gamma \, \mathbb{E}_{p_{s,a}(X), X\sim \mathbb{P}_g}\left[\max_{a'} \tilde{Q}(s', a') \right]\right\},
\end{equation}
which provides an upper bound for the distributionally robust Bellman operator, as shown by:
\begin{equation}
    \begin{aligned}
        \tilde{\mathcal{U}}_{\mathfrak{R}}(\tilde{Q})(s, a) & \geq \inf_{g \in \mathcal{G}} \left\{ \mathbb{E}_{X \sim \mathbb{P}_g} [r(s, a;X)] \right\} + \inf_{g \in \mathcal{G}} \left\{  \gamma \,  \mathbb{E}_{p_{s,a}(X), X\sim \mathbb{P}_g}\left[\max_{a'} \tilde{Q}(s', a') \right] \right\}\\
        & \geq \inf_{g \in \mathcal{G}} \left\{ \mathbb{E}_{X \sim \mathbb{P}_g} [r(s, a;X)] \right\} + \inf_{\mathbb{P} \in \mathfrak{M}} \left\{\gamma \, \mathbb{E}_{p_{s,a}(X), X\sim \mathbb{P}_g}\left[\max_{a'} \tilde{Q}(s', a')  \right] \right\}\\
        & = \tilde{\mathcal{T}}_{\mathfrak{R}}(\tilde{Q})(s, a).
    \end{aligned}
\end{equation}

During training, we use \eqref{eq:dr_bo_approx} to construct the loss function \eqref{eq:drcb_loss} for DRMARL, where the optimization problem within \eqref{eq:dr_bo_approx} is solved using the solution from the CB-based worst-case reward predictor $\qcb$. In practice, this approximation \eqref{eq:dr_bo_approx} proves highly effective for the worst-case return, with the relative approximation error of $\tilde{\mathcal{U}}_{\mathfrak{R}}(\tilde{Q})(s, a)$ to $\tilde{\mathcal{T}}_{\mathfrak{R}}(\tilde{Q})(s, a)$ being less than $0.57\%$ (see \cref{fig:trans_prob_compare}). This small error margin indicates that $\tilde{\mathcal{U}}_{\mathfrak{R}}(\tilde{Q})(s, a)$ does not impede DRMARL's ability to observe the worst-case return.

\section{Additional Simulation Result for Large-Scale Amazon Robotic Sortation Environments} \label{app:detail_validation}



\begin{table}[t]
\caption{Key metrics improvements over MARL baseline trained on Year 4 data in large-scale Amazon robotic sortation warehouses.}
\label{tab:end_metric_all}
\vskip 0.15in
\begin{center}
\begin{small}
\begin{sc}
\begin{tabular}{ccccc}
\toprule
\makecell{\textbf{Group} \\\textbf{Number}} & \textbf{Year} & \makecell{\textbf{Recirculation Rate} \\\textbf{Reduction (\%)} } & \makecell{\textbf{Package Throughput} \\\textbf{Increase (\%)}} & \makecell{\textbf{Package Recirculation} \\ \textbf{Amount Reduction (\%)}} \\
\midrule
1 & 1 & 94.75 & 31.66 & 94.21 \\
2 & 1 & 93.19 & 23.47 & 92.55 \\
3 & 1 & 94.62 & 35.69 & 94.26 \\
4 & 1 & 93.85 & 25.94 & 93.85 \\
5 & 1 & 95.90 & 19.50 & 95.67 \\
6 & 2 & 90.47 & 5.79 & 90.44 \\
7 & 2 & 91.27 & 5.24 & 91.11 \\
8 & 2 & 77.22 & 4.13 & 77.22 \\
9 & 2 & 83.34 & 2.49 & 83.34 \\
10 & 2 & 85.12 & 5.07 & 85.12 \\
11 & 2 & 92.85 & 6.97 & 92.82 \\
12 & 3 & 84.22 & 4.54 & 84.59 \\
13 & 3 & 83.14 & 15.89 & 81.57 \\
14 & 3 & 85.53 & 4.17 & 86.04 \\
15 & 3 & 83.63 & 5.51 & 83.58 \\
16 & 3 & 84.24 & 5.52 & 84.19 \\
17 & 4 & 34.18 & -0.99 & 36.71 \\
18 & 4 & 57.04 & -5.85 & 62.34 \\
19 & 4 & 75.78 & 4.47 & 75.73 \\
20 & 4 & 66.59 & 9.85 & 64.18 \\
21 & 4 & 75.07 & 5.64 & 75.83 \\
\bottomrule
\end{tabular}
\end{sc}
\end{small}
\end{center}
\vskip -0.1in
\end{table}


\cref{tab:end_metric_all} presents detailed validation results comparing MARL and DRMARL chute mapping policies across all induction distribution groups from Year 1-4. The DRMARL policy demonstrates superior performance across most groups, achieving both higher package sortation and lower recirculation rates. The only exceptions are two groups in Year 4, where the MARL policy shows marginally better throughput but at the cost of higher recirculation rates. This is expected behavior since the MARL policy is specifically trained on Year 4 induction data, while DRMARL optimizes for robustness rather than throughput maximization. Overall, DRMARL achieves significant improvements, reducing recirculation by $80\%$ on average while simultaneously increasing throughput by $5.62\%$ on average.

\end{document}